\makeatletter\def\graphicscache@inhibit{true}\makeatother

\documentclass[a4paper, 10pt, conference]{ieeeconf}      %

\IEEEoverridecommandlockouts                              %

\usepackage{cite}
\usepackage{amsmath,amssymb,amsfonts}
\usepackage{algorithmic}
\usepackage{graphicx}
\usepackage{textcomp}
\usepackage{xcolor}
\definecolor{apple}{HTML}{4CC417}
\definecolor{amber}{HTML}{FFBF00}
\definecolor{airblue}{HTML}{77ACC7}
\definecolor{darkgreen}{HTML}{006400}
\definecolor{lila}{HTML}{E0B0FF}

\usepackage{makeidx}

\usepackage{pgfplots}
\pgfplotsset{compat=newest}
\usepackage{multirow}
\usepackage{tabularx}
\usepackage{threeparttable}
\usepackage{booktabs}

    \newcommand{\midsepremove}{\aboverulesep = 0mm \belowrulesep = 0mm}
    \midsepremove
    \newcommand{\midsepdefault}{\aboverulesep = 0.4ex \belowrulesep = 0.65ex}
    \midsepdefault
    
\usepackage{array}
\usepackage{float}
\usepackage{graphbox}
\usepackage{tikz,tikz-3dplot}
\usetikzlibrary{fit,arrows,arrows.meta,automata,backgrounds,calc,chains,%
decorations.markings,decorations.pathreplacing,decorations.pathmorphing,%
matrix,positioning,shapes,shapes.geometric,shapes.symbols,spy,trees,tikzmark}
\usepackage{balance}
\usepackage[binary-units=true,product-units=single,per-mode=symbol,range-units=single,range-phrase=\,--\,,detect-all]{siunitx}
\DeclareSIUnit\pixel{px}
\usepackage{bm}
\usepackage{acronym}
\usepackage{tablefootnote}
\usepackage{hyperref}

\let\labelindent\relax
\usepackage[inline]{enumitem}

\definecolor{bg_color}{RGB}{95,95,95}

\newcommand{\reffig}[1]{Fig.~\ref{#1}}
\newcommand{\reftab}[1]{Tab.~\ref{#1}}
\newcommand{\refsec}[1]{Sec.~\ref{#1}}

\usepackage{adjustbox}
\newcolumntype{R}[2]{%
    >{\adjustbox{angle=#1,lap=\width-(#2)}\bgroup}%
    l%
    <{\egroup}%
}
\newcolumntype{L}[1]{>{\raggedright\let\newline\\\arraybackslash\hspace{0pt}}m{#1}}

\title{\LARGE \bf
Marker-free Human Gait Analysis using a Smart Edge Sensor System
}

\author{Eva Katharina Bauer$^{1}$, Simon Bultmann$^{2}$, and Sven Behnke$^{2}$%
\thanks{This work was supported by WestAI, Grant No. 01IS22094A.}%
\thanks{$^{1}$Hochschule Koblenz, RheinAhrCampus, Remagen, Germany;
        {\tt\small \{ebauer1\}@hs-koblenz.de,}}%
\thanks{$^{2}$Autonomous Intelligent Systems group, %
		University of Bonn, Germany;
        {\tt\small \{bultmann,behnke\}@cs.uni-bonn.de}}%
}

\begin{document}

\maketitle
\thispagestyle{empty}
\pagestyle{empty}

\begin{abstract}
The human gait is a complex interplay between the neuronal and the muscular systems, reflecting an individual's neurological and physiological condition. This makes gait analysis a valuable tool for biomechanics and medical experts. Traditional observational gait analysis is cost-effective but lacks reliability and accuracy, while instrumented gait analysis, particularly using marker-based optical systems, provides accurate data but is expensive and time-consuming.\\
In this paper, we introduce a novel markerless approach for gait analysis using a multi-camera setup with smart edge sensors to estimate 3D body poses without fiducial markers. We propose a Siamese embedding network with triplet loss calculation to identify individuals by their gait pattern. This network effectively maps gait sequences to an embedding space that enables clustering sequences from the same individual or activity closely together while separating those of different ones.
Our results demonstrate the potential of the proposed system for efficient automated gait analysis in diverse real-world environments, facilitating a wide range of applications. 
\end{abstract}

\section{Introduction}
\label{sec:Introduction}
Locomotion, particularly walking, is an essential ability for humans. It is learned at an early age and is usually a subconscious process. Moving on two legs gives us independence and makes physical activities like running possible.

Because of its complexity and uniqueness, gait analysis attracts significant interest from both experts in biomechanics and medical professionals. Observing and analyzing changes in gait can not only provide insight into neurological and physiological conditions, but can also aid in the development and evaluation of individualized treatments~\cite{whittle1996clinical, hulleck2022present}.

The clinical use of gait analysis mainly focuses on observational gait analysis performed with human eye and brain. This method is simple and cost-efficient but suffers from low validity, reliability, and responsiveness. Instrumented gait analysis on the other hand promises accurate and reliable gait data for medical use~\cite{hulleck2022present}. Marker-based optical motion capture systems are considered the gold standard in instrumented gait analysis. They provide a high level of precision, but are expensive and time-consuming to use~\cite{moro2022markerless}. 

\begin{figure}[t]
	\centering
	\resizebox{1.\linewidth}{!}{
		\begin{tikzpicture}
		    \node[inner sep=0,anchor=north east] (image1) at (0, 0){\includegraphics[height=2.5cm,trim= 0 0 0 0, clip]{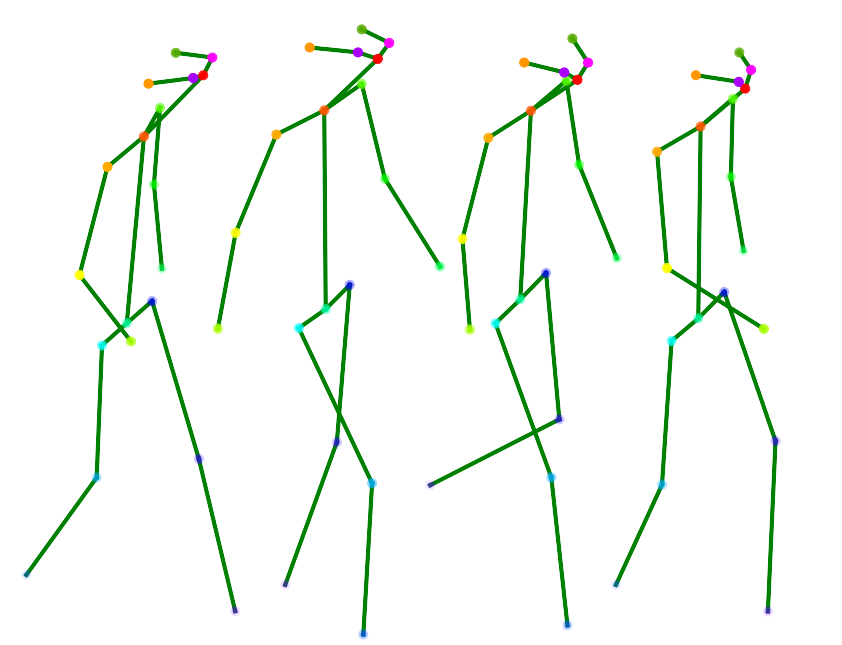}};
		    \node[inner sep=0,anchor=south,xshift=0., yshift=0.3cm] (image2) at (image1.north) {\includegraphics[height=1.cm,trim= 0px 600px 0px 350px, clip]{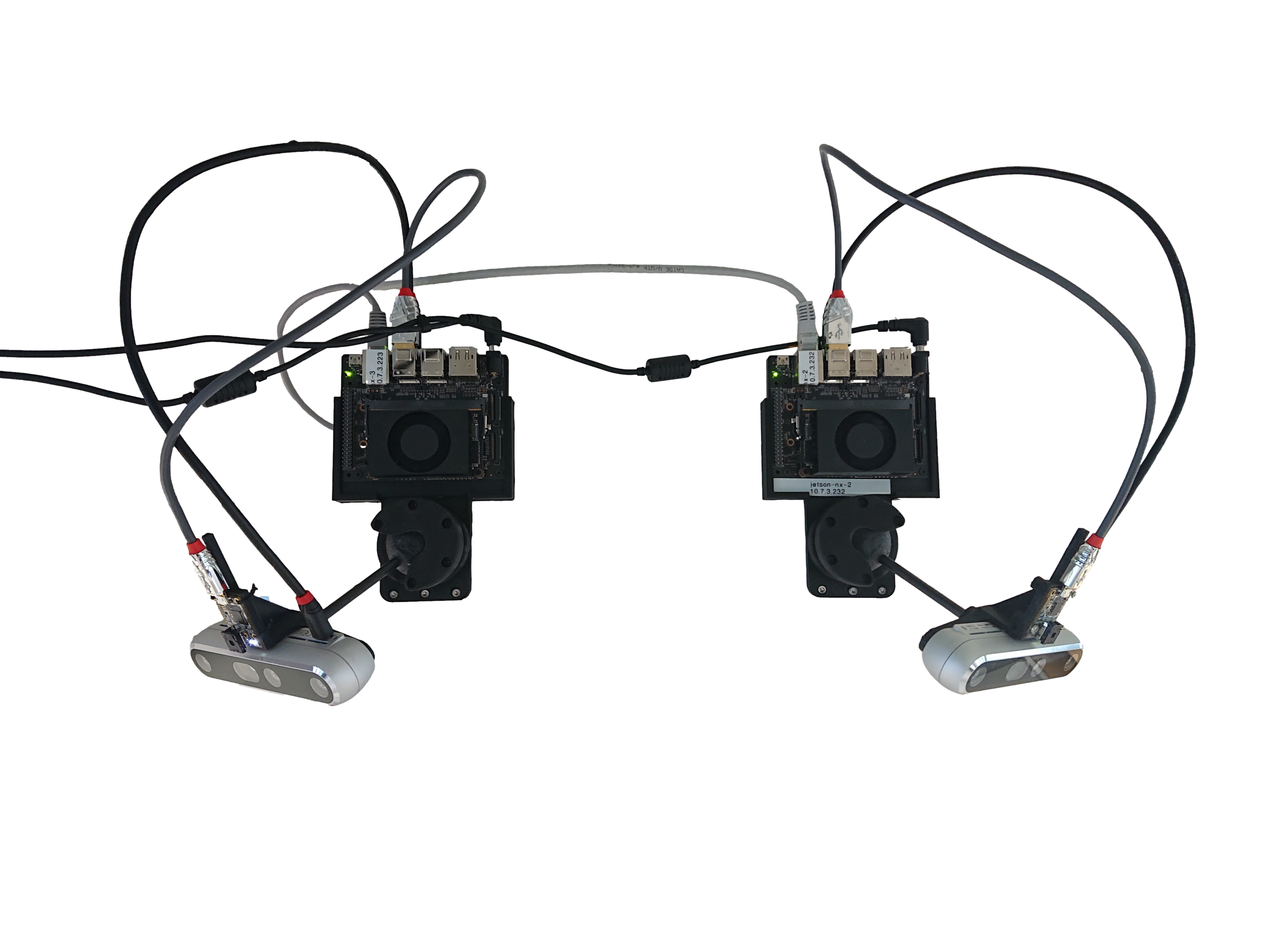}};
		    
		    \node[inner sep=0.2,anchor=south west, yshift=-0.2cm, xshift=0.2cm, draw=gray, thick, rounded corners] (image3) at (image1.south east){\includegraphics[height=4.cm,trim= 0 0 0 0, clip]{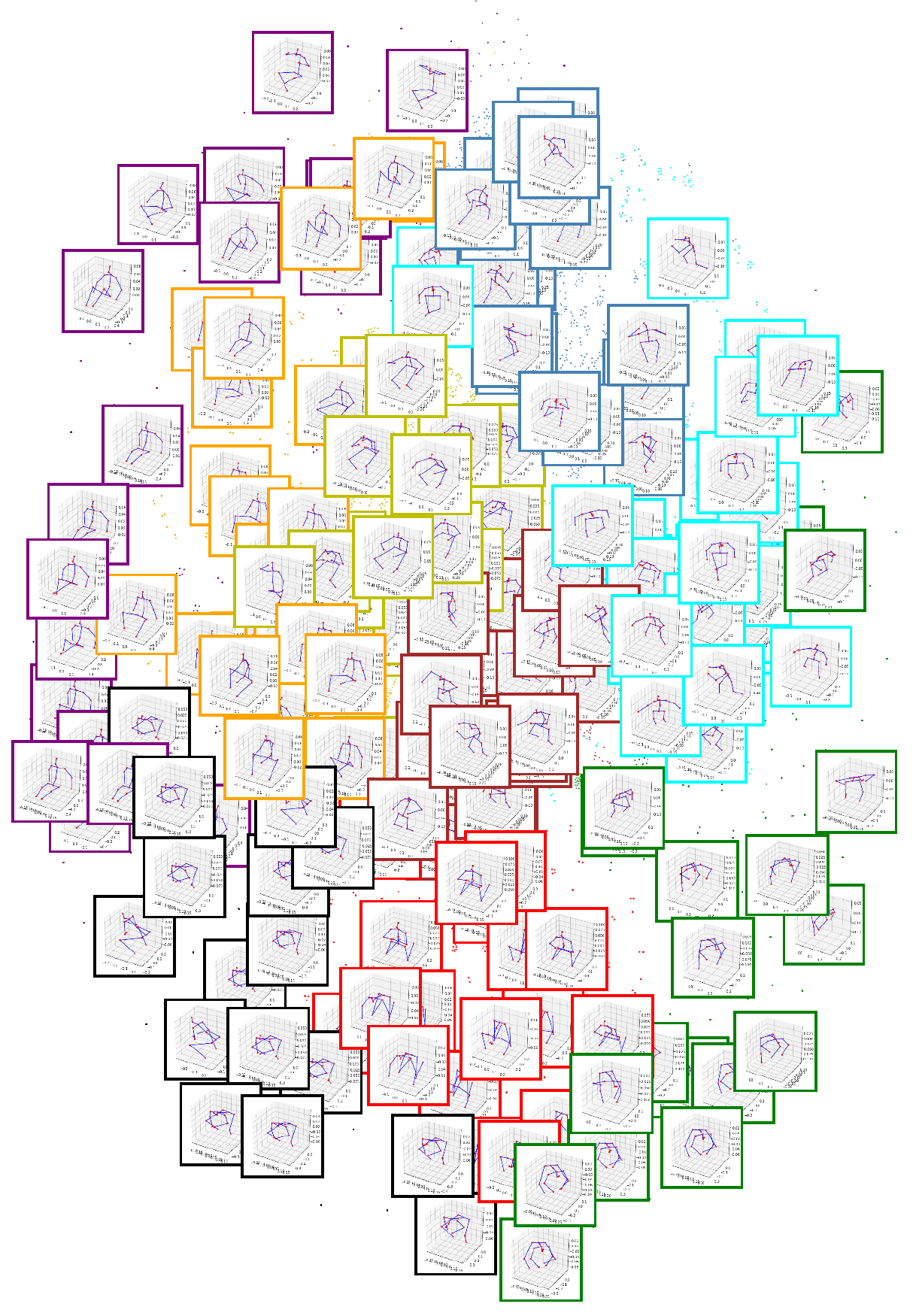}};
		    \node[inner sep=0,anchor=west, xshift=0.05cm] (image4) at (image3.east){\includegraphics[height=3.cm,trim= 0 0 0 0, clip]{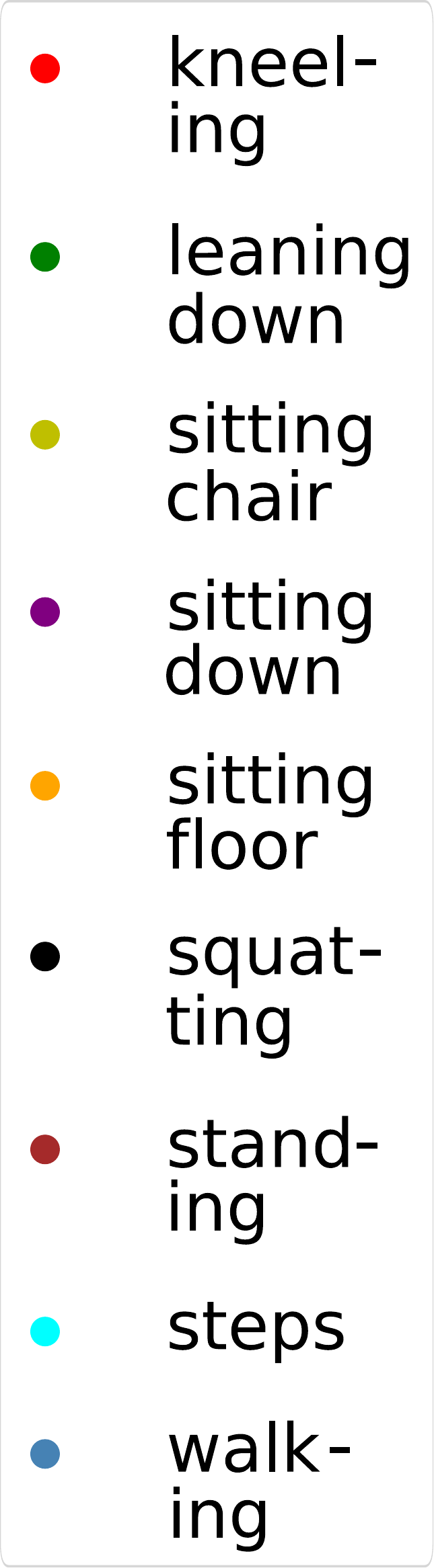}};

		    \draw[->, thick] (image1.south west) -- (image1.south east) node[near end, below, xshift=.4cm, font=\scriptsize]{Time};
		    
			\node[label,scale=1., anchor=south west,yshift=-0.05cm,xshift=-0.6cm, rectangle, fill=white, align=center, font=\scriptsize\sffamily] (n_1) at (image2.south west) {(a)};		    
		    \node[label,scale=1., anchor=south west,yshift=0.03cm,xshift=-0.17cm, rectangle, fill=white, align=center, font=\scriptsize\sffamily] (n_0) at (image1.south west) {(b)};
		    \node[label,scale=1., anchor=south west,yshift=0.03cm,xshift=0.03cm, rectangle, align=center, font=\scriptsize\sffamily] (n_2) at (image3.south west) {(c)};

\end{tikzpicture}
}
\vspace{-2.em}
    \caption{Sample movement sequence~(b) captured with smart edge sensors~(a) consisting of a Nvidia Jetson Orin compute board and an Intel RealSense RGB-D camera. (c) t-SNE visualization of activities from Human~3.6M dataset based on the learned gait embedding.}
    \label{fig:Titel}
    \vspace{-1.em}
\end{figure}
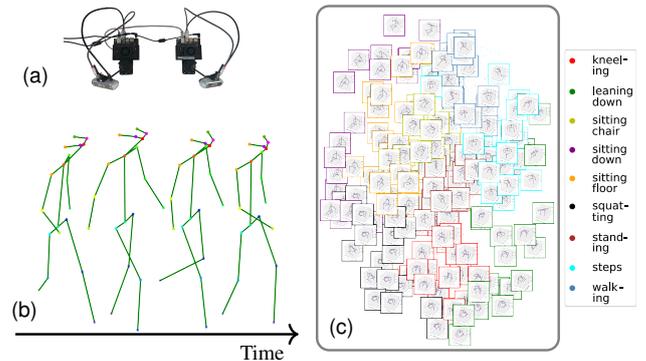

In this work, we instead employ a markerless approach to capture human movement, developed in previous work~\cite{Bultmann_RSS_2021, bultmann_ias2022, paetzold_camcalib_2022}. A multi-camera system is used to estimate the 3D poses of multiple persons in real time, as illustrated in \reffig{fig:Titel}\,(a, b).
We refer to the camera nodes as \textit{smart edge sensors}, as they contain an integrated inference accelerator for local, on-board, semantic image interpretation.
Unlike marker-based systems that require applying markers to the participant's body (e.g. wearing a marker suit), the smart edge sensors detect human joints by inferring heatmaps of human body keypoints from the camera images. 
These keypoint detections are streamed to a central backend, where the pose estimates of each camera are fused into a 3D skeleton per observed person. The fusion process incorporates priors on typical bone lengths to enhance the accuracy of the pose estimation~\cite{Bultmann_RSS_2021, bultmann_ias2022}. The proposed system enables the capture of human motion sequences and gait measurements in real time, making gait analysis accessible for wider use in diverse, real-world environments.

Another significant challenge in clinical gait analysis is the wide range of parameters affecting human gait. Efficiently capturing gait patterns and accurately recognizing individuals based on them could simplify the differentiation between patient groups. Identifying similar gait characteristics across different patients may indicate a shared underlying condition, aiding in the selection of appropriate treatments. Integrating machine learning methods into this analysis facilitates recognizing individual gait patterns and clustering similar traits or activities across different individuals \cite{slemenvsek2023human, khera2020role, hummel2024clustering,horst2019explaining}.

In this work, we design a Siamese embedding network based on TriNet~\cite{hermans2017triplet} for the identification of individuals by their walking patterns and for the differentiation of activities. The network takes gait sequences as input and maps them into an embedding space, using a ResNet\,18 backbone~\cite{he_deep_2016}. We train the network using the Triplet Loss~\cite{weinberger2009distance} on $L_2$-distances in the embedding space to ensure that motion sequences from the same person or activity are positioned close together, while sequences from different individuals or actions are pushed apart (cf. \reffig{fig:Titel}~(c)).

In summary, our main contributions in this paper are:
\begin{itemize}
	\item We propose a novel deep learning-based framework to cluster human walking patterns in an embedding space to identify similar gait patterns or activities.
    \item We collect accurate gait data from multiple subjects in a real-world environment using a smart edge sensor network to capture human movement without the need to apply markers or sensors to the body.
    \item We quantitatively and qualitatively evaluate the proposed integrated system for human motion capture and gait analysis using the collected real-world data and the Human~3.6M database~\cite{h36m}.
\end{itemize}

\section{Related Work}
\label{sec:Related_Work}
The development of approaches for human gait pattern analysis with artificial neural networks started in 1993 with Holzreiter et al.~\cite{holzreiter1993assessment} proposing a three-layer neural network with connected units to distinguish between healthy and pathological gait patterns using data captured by force measurement platforms.
In 2002, Sch\"ollhorn et al.~\cite{schollhorn2002identification} proposed to determine individual movement characteristics using self-organizing maps and data from force platforms. They compared the performance using time-continuous and time-discrete data concluding that continuous data leads to more accurate and stable results.
In 2006 Han et al.~\cite{han2005individual} introduced the \textit{gait energy image} (GEI).
Instead of using a sequence of templates they fuse human motion in a single image using component and discriminant analysis to learn gait features from the images.

Supervised machine learning approaches were introduced in 2007 by Lu et al.~\cite{lu2007gait}. Their approaches are based on Independent Component Analysis (ICA) and Principal Component Analysis (PCA). As input, images of moving persons are reduced to binary silhouettes by background subtraction and then characterized by mathematical methods. Although their recognition accuracy was promising, the method cannot be reliably used for automatic human identification in real-world environments. Main limitations are the unpredictability of the angle between the walker and the camera and the lack of a more varied gait database.

Alotaibi et al.~\cite{alotaibi2017improved} proposed a specialized deep Convolutional Neural Network (CNN) architecture for gait recognition. They used the CASIA-B dataset~\cite{yu2006framework} and an eight-layer CNN to recognize individuals by their gait. In our work, we employ the widely used and efficient ResNet~\cite{he_deep_2016} backbone instead.
In 2023, Taha et al.~\cite{taha2024learning} introduced an auto-encoder to recognize biological and physiological characteristics of individuals, like gender, age, and weight. They used Inertial Measurement Units (IMUs) located in both outsoles of the shoes as well as a marker-based motion capture system with infrared cameras to collect gait data of people walking on a treadmill. After the learning process, they clustered the network output using the K-Means algorithm. Their clustering-based model achieved a high identification accuracy. In our work, in contrast, we employ a marker-free optical capture setup and need to neither add sensors nor markers to the subjects for the gait analysis, making the capture process significantly more time-efficient and accessible.

Schroff et al.~\cite{schroff2015facenet} also attempt to recognize and cluster people by specific traits. They recognize the faces of people by reducing the input data to an embedding vector and clustering embeddings of pictures of the same person. They propose the semi-hard mining process for the triplet loss calculation and achieve accurate clustering, benefiting from their large dataset. We take up the semi-hard mining strategy in our training process but process the more privacy-preserving gait sequence data, as all image processing happens locally on the smart edge sensors in our system. 

\section{Method}
\label{sec:Method}
\begin{figure}[t]
	\centering \footnotesize 
	\begin{tabular}{ccc}
		\includegraphics[height=4cm]{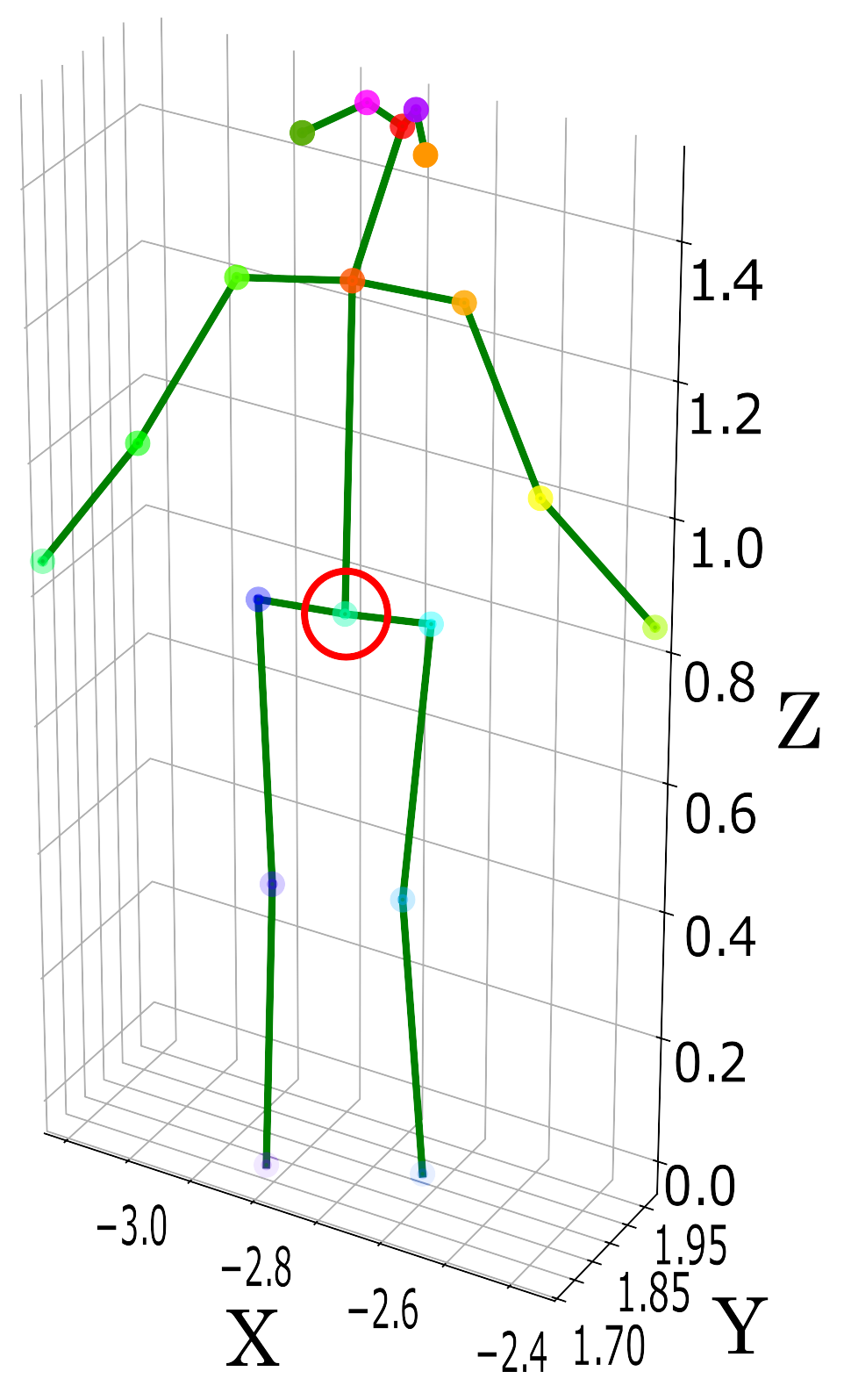} &
		\includegraphics[height=4cm]{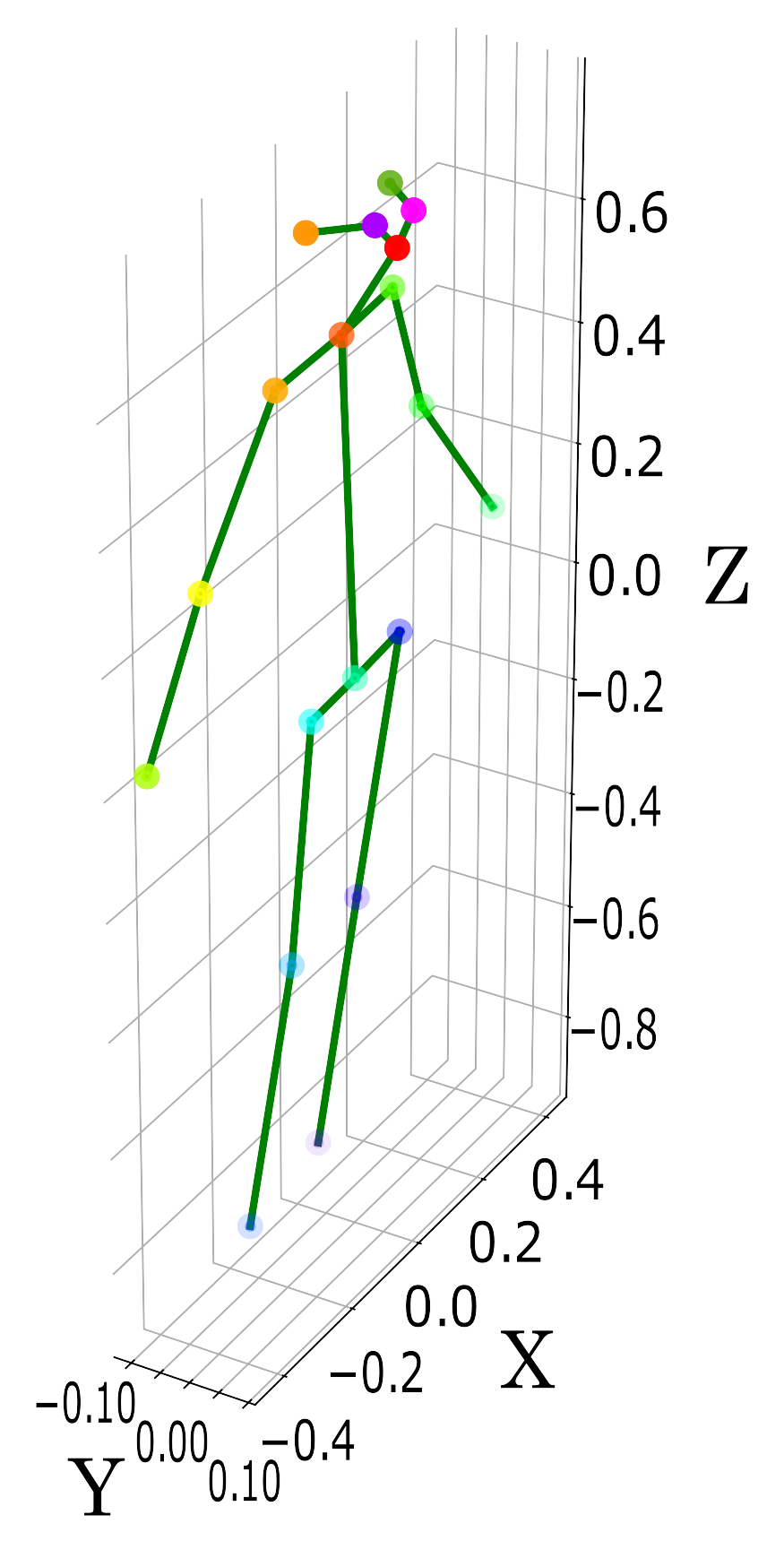} &
		\includegraphics[height=4cm]{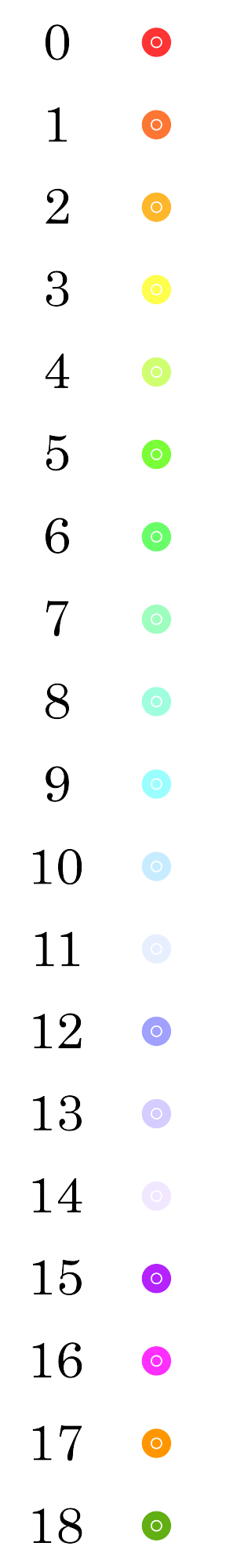}\\
		\textbf{(a)} 3D skeleton &
		\textbf{(b)} Normalized skeleton & \\[.5em]
		\multicolumn{3}{c}{\includegraphics[width=.7\linewidth]{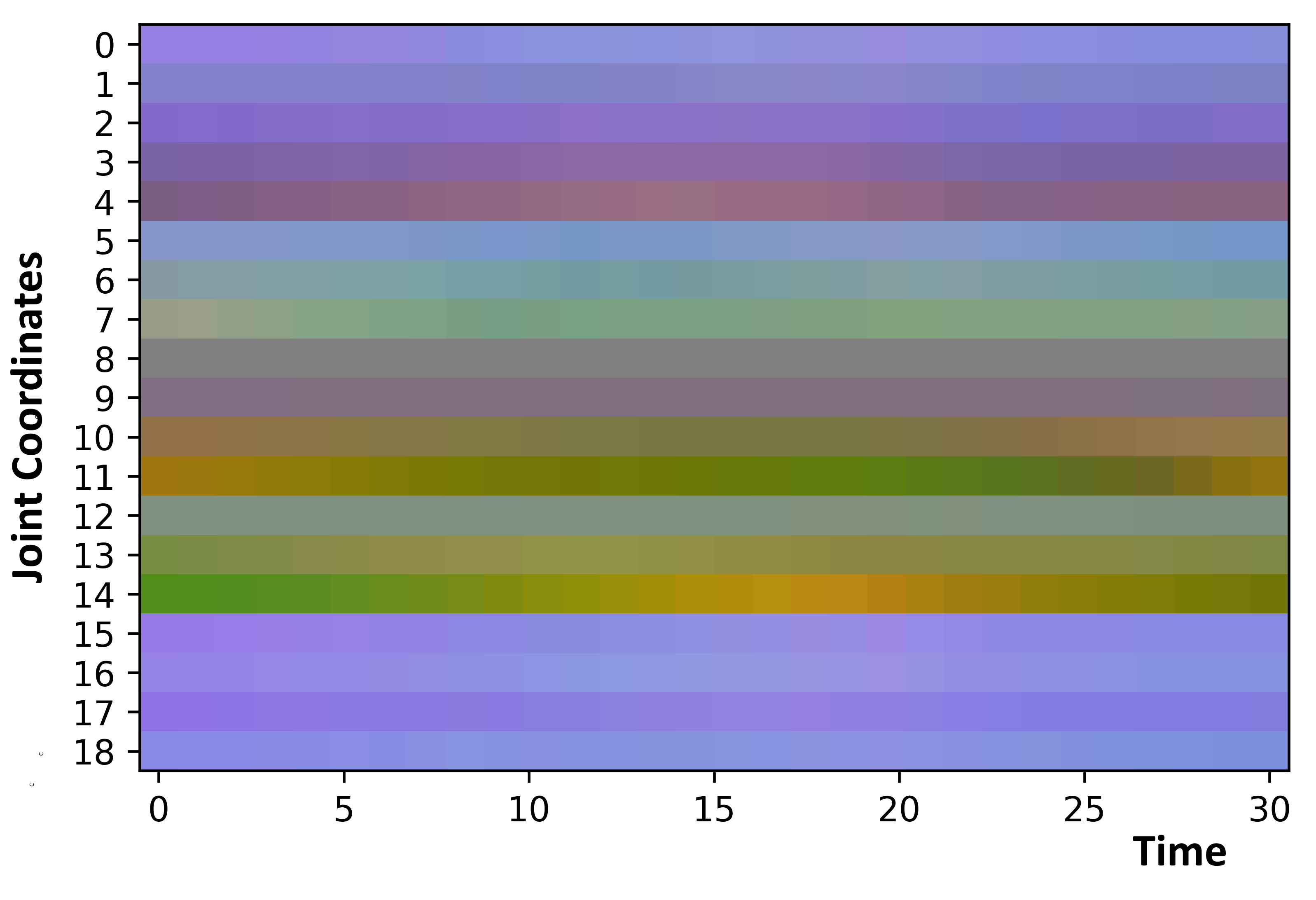}}\\[-.7em]
		\multicolumn{3}{c}{\textbf{(c)} Input tensor}
	\end{tabular}
  
    \caption{Example of the 3D skeleton model with $J=19$ joints~(a), coded by color. The red circle marks the root joint used for normalization (b); (c) RGB-representation of one $(J,T,3)$ input tensor with $T=30$ frames.}
    \label{fig:3DModel}
    \vspace{-1em}
\end{figure}
In this section, we first describe the smart edge sensor system used for gait data collection. We then detail the data-prepossessing steps and introduce the proposed neural network designed to cluster different gait patterns.

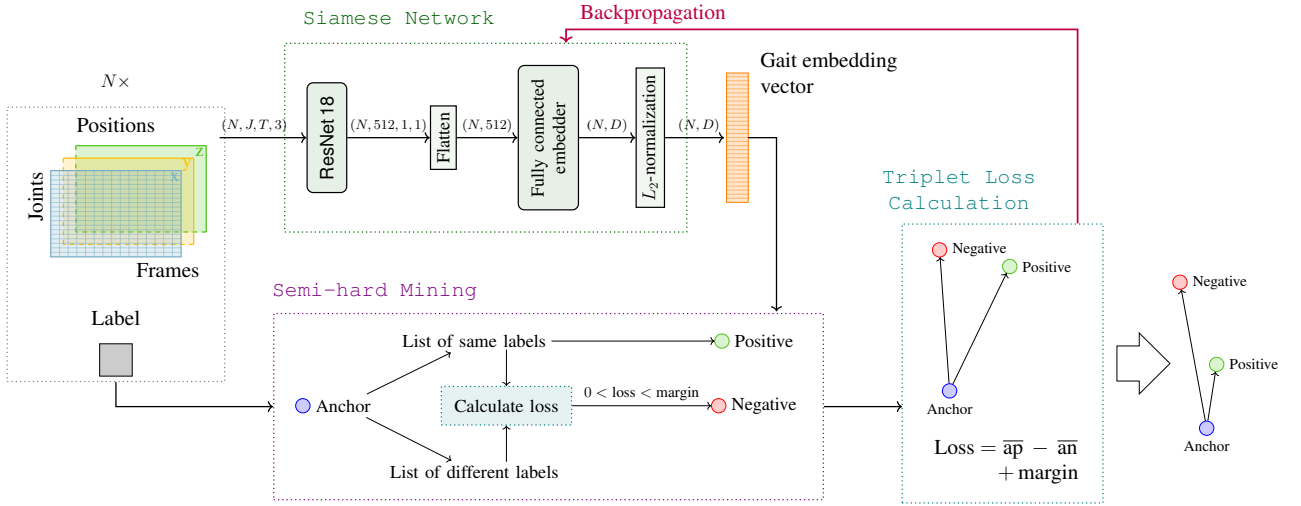
\begin{figure*}[t]
    \centering
    \resizebox{1.\textwidth}{!}{
            \newsavebox{\inbox}
            \sbox{\inbox}{
                \begin{tikzpicture}
                    \coordinate (I) at (0,2,-1.5);
                    \coordinate (J) at (3,2,-1.5);
                    \coordinate (K) at (3,0,-1.5);
                    \coordinate (L) at (0,0,-1.5);

                    \draw[fill, opacity=0.2, apple] (I) -- (J) -- (K) -- (L) -- cycle;
                    \draw[thick, apple] (I) -- (J);
                    \draw[thick, apple] (J) -- (K);
                    \draw[thick, dashed, apple] (K) -- (L);
                    \draw[thick, dashed, apple] (L) -- (I);

                    \node at (J) [yshift=-0.15cm, xshift=-0.15cm, font=\large, apple] {z};        

                    \coordinate (E) at (0,2,-0.75);
                    \coordinate (F) at (3,2,-0.75);
                    \coordinate (G) at (3,0,-0.75);
                    \coordinate (H) at (0,0,-0.75);
                    \draw[fill, opacity=0.2, amber] (E) -- (F) -- (G) -- (H) -- cycle;
                    \draw[thick, amber] (E) -- (F);
                    \draw[thick, amber] (F) -- (G);
                    \draw[thick, dashed, amber] (G) -- (H);
                    \draw[thick, dashed, amber] (H) -- (E);

                    \node at (F) [yshift=-0.15cm, xshift=-0.15cm, font=\large, amber] {y};            

                    \coordinate (A) at (0,0,0);
                    \coordinate (B) at (3,0,0);
                    \coordinate (C) at (3,2,0);
                    \coordinate (D) at (0,2,0);
                    \draw[thick, airblue, fill, opacity=0.2, draw opacity=1] (A) -- (B) -- (C) -- (D) -- cycle;

                    \node at (C) [yshift=-0.15cm, xshift=-0.15cm, font=\large, airblue] {x};

                    \foreach \x in {1,...,14}{
                        \coordinate (X) at (\x*0.2,0,0);
                        \coordinate (Y) at (\x*0.2,2,0);
                        \draw [thin, airblue, draw opacity=0.5] (X) -- (Y);
                    }

                    \foreach \x in {1,...,18}{
                        \coordinate (X) at (0,0.105*\x,0);
                        \coordinate (Y) at (3,0.105*\x,0);
                        \draw [thin, airblue, draw opacity=0.5] (X) -- (Y);
                    }

                    \node at (D) [rotate=90, yshift=+0.35cm, xshift=-0.6cm,font=\Large] {Joints};
                    \node at (B) [yshift=-0.3cm, xshift=-0.3cm,font=\Large] {Frames};
                \end{tikzpicture}
            }

            \newsavebox{\outbox}
            \sbox{\outbox}{
                \begin{tikzpicture}
                    \coordinate (A) at (0, 0, 0);
                    \coordinate (B) at (0, 3, 0);
                    \coordinate (C) at (0.5, 3, 0);
                    \coordinate (D) at (0.5, 0, 0);

                    \draw[fill, opacity=0.2, orange, draw opacity=1] (A) -- (B) -- (C) -- (D) -- cycle;

                    \foreach \x in {1,...,31}{
                        \coordinate (X) at (0,0.094*\x,0);
                        \coordinate (Y) at (0.5,0.094*\x,0);
                        \draw [thin, orange, draw opacity=0.5] (X) -- (Y);
                    }
                    
                \end{tikzpicture}
            }

            \newsavebox{\lossBox}
            \sbox{\lossBox}{
                \begin{tikzpicture}
                    \tikzstyle{dott} = [circle, opacity=0.2, draw opacity=1]

                    \coordinate (A) at (0,0,0);
                    \coordinate (B) at (1., 2.5,-1);
                    \coordinate (C) at (-1.,2.5,-2);
                    
                    \node (anc) at (A) [dott, draw=blue, fill=blue]{};
                    \node [below =of anc, yshift=1cm]{Anchor};
                    \node (pos) at (B) [dott, draw=apple, fill=apple]{};
                    \node [right =of pos, xshift=-1cm]{Positive};
                    \node (neg) at (C) [dott, draw=red, fill=red]{};
                    \node [right =of neg, xshift=-1cm]{Negative};

                    \draw [->] (anc) -- (pos);
                    \draw [->] (anc) -- (neg);

                    \node (formular)[align = center, below =of A, yshift=0.cm, xshift=1.3cm, font=\Large]{Loss  $ = \overline{\text{ap}}\,-\, \overline{\text{an}}$\\$\qquad\quad+\,\text{margin}$};
                \end{tikzpicture}
            }

            \newsavebox{\tripletBox}
            \sbox{\tripletBox}{
                \begin{tikzpicture}
                    \tikzstyle{dott} = [circle, opacity=0.2, draw opacity=1]

                    \node (anc) [dott, draw=blue, fill=blue]{};
                    \node (ancL) [right =of anc, xshift=-1cm, font=\large]{Anchor};

                    \node (same) [right =of anc, yshift=1.5cm, xshift=1.cm, font=\large] {List of same labels};
                    \node (dif) [below =of same, yshift= -1.5cm, font=\large] {List of different labels};

                    \node (pos) [right =of same, xshift=2.75cm, dott, draw=apple, fill=apple]{};
                    \node (posL) [right =of pos, xshift=-1cm, font=\large]{Positive};

                    \draw [->] (ancL) -- (same);
                    \draw [->] (ancL) -- (dif);
                    
                    \draw [->] (same) -- (pos);

                    \node (loss) [rectangle, inner sep=.3cm, dotted, thick, right =of anc, xshift=2cm, draw=teal, fill=teal!10, font=\large]{\textcolor{black}{Calculate loss}};
                    \draw [->] (loss) ++(0.cm, 1.3cm) -- (loss);
                    \draw [->] (loss) ++(0.cm, -1.3cm) -- (loss);

                    \node (neg) [right =of loss, dott, xshift=2.2cm, draw=red, fill=red]{};
                    \node (negL) [right =of neg, xshift=-1.cm, font=\large]{Negative};

                    \draw [->] (loss) -- (neg) node[midway, above]{$0 < \text{loss} < \text{margin}$};
                    
                \end{tikzpicture}
            }

            \newsavebox{\resBox}
            \sbox{\resBox}{
                \begin{tikzpicture}
                    \tikzstyle{dott} = [circle, opacity=0.2, draw opacity=1]

                    \coordinate (A) at (0,0,0);
                    \coordinate (B) at (1., 2.25,2);
                    \coordinate (C) at (-1, 3,-1);
                    
                    \node (anc) at (A) [dott, draw=blue, fill=blue]{};
                    \node [below =of anc, yshift=1cm]{Anchor};
                    \node (pos) at (B) [dott, draw=apple, fill=apple]{};
                    \node [right =of pos, xshift=-1cm]{Positive};
                    \node (neg) at (C) [dott, draw=red, fill=red]{};
                    \node [right =of neg, xshift=-1cm]{Negative};

                    \draw [->] (anc) -- (pos);
                    \draw [->] (anc) -- (neg);

                \end{tikzpicture}
            }
            
        \begin{tikzpicture}[auto]
            \node (posP) [] {\usebox{\inbox}};
            \node (pos)  [above =of posP, yshift=-1cm,font=\Large] {Positions};

            \node (lblP) [below =of posP, yshift=-.3cm, rectangle, minimum width=0.75cm, minimum height=0.75cm, draw, fill, fill opacity=0.2]{};
            \node (lbl)  [above =of lblP, yshift=-0.7cm, font=\Large]{Label};

            \node (input) [fit=(posP)(pos)(lblP)(lbl), thick, dotted, gray, draw]{};
            
            \node [above =of input, yshift=-.7cm, font=\large]{$N \times$};

            \node (rn)          [rectangle, rounded corners, rotate=90, inner sep=.3cm, draw=black, fill=darkgreen!10, thick, right =of posP.north east, anchor=north, yshift=-1.cm, text=black, font=\large]{\sf ResNet\,18};
            \node (flatt)       [rectangle, rotate=90, inner sep=.15cm, draw=black, thick, fill=darkgreen!5, right =of rn.south, anchor=north, yshift=-.9cm, font=\large]{Flatten};
            \node (fce)         [rectangle, rounded corners, rotate=90, inner sep=.3cm, draw=black, thick, fill=darkgreen!10, right =of flatt.south, anchor=north, yshift=-.4cm, font=\large, align=center]{Fully connected\\embedder};
            \node (norm)        [rectangle, rotate=90, inner sep=.15cm, draw=black, thick, fill=darkgreen!5, right =of fce.south, anchor=north, font=\large, yshift=-.3cm]{$L_2$-normalization};

            \node (siamese)    [fit=(rn)(fce)(norm), draw=darkgreen, dotted, thick, inner sep=0.5cm ] {};
            \node                 [above =of siamese, yshift=-.7cm, xshift=-2cm, font=\Large] {\textcolor{darkgreen}{\texttt{Siamese Network}}};
            
  	        \draw [->, line width=.3mm] (posP.north east) -- (rn.north) node[midway, above, font=\small, xshift=-.2cm]{$(N, J, T, 3)$};
            \draw [->, line width=.3mm] (rn) -- (flatt) node[midway, above, font=\small, xshift=0cm]{$(N, 512, 1, 1)$};
            \draw [->, line width=.3mm] (flatt) -- (fce) node[midway, above, font=\small, xshift=0.cm]{$(N,512)$};
            \draw [->, line width=.3mm] (fce) -- (norm) node[midway, above, font=\small, xshift=0.cm]{($N, D)$};

            \node (emb) [right =of norm.center, xshift=0.5cm]{\usebox{\outbox}};
            \node (emb2) [right =of norm.center, xshift=0.7cm]{};
            \node (emb3) [right =of norm.center, xshift=1.0cm]{};
            \node [align = left, right =of emb, yshift=1.5cm, xshift=-1.1cm, font=\Large]{Gait embedding\\vector};
            \draw [->, line width=.3mm] (norm) -- (emb2) node[midway, above, font=\small, xshift=0.1cm]{$(N, D)$};

            \node (mining) [right =of lblP.south east, yshift=-.7cm, xshift=2.5cm]{\usebox{\tripletBox}};
            \node (min) [fit=(mining), draw=violet, dotted, thick, inner sep=0.25cm ]{};
            \node       [above =of min, yshift=-.8cm, xshift=-4.cm, font=\Large] {\textcolor{violet}{\texttt{Semi-hard Mining}}};

            \node (a) [below =of emb, xshift=.9cm, yshift=-1.4cm]{};
            \draw [->, line width=.3mm] (emb3) -| (a);
            \draw [->, line width=.3mm] (input) |- (min);

            \node (lossDist) [right =of min, xshift=1cm, yshift=1.cm] {\usebox{\lossBox}};
            \node (lossD) [fit=(lossDist), draw=teal, dotted, thick, inner sep=0.2cm] {};
            \node         [align=center, above =of lossD, yshift=-.8cm, xshift=-1.cm, font=\Large] {\textcolor{teal}{\texttt{Triplet Loss}}\\\textcolor{teal}{\texttt{Calculation}}};

            \node (b) [right =of min, xshift=0.8cm]{};
            \draw [->, line width=.3mm] (min) -- (b);

            \draw [->, purple, line width=.4mm] (lossD.62) |- ++(0, 4.5cm) -| (siamese.50) node[midway, above, xshift=2cm, font=\Large]{Backpropagation};

           \node [right =of lossD, xshift=-0.7cm, yshift=0cm, single arrow,draw,inner sep=0.4cm, minimum height=.3cm, single arrow head extend=0.3cm] (resolution) {};

            \node [right =of resolution, xshift=-1.2cm]{\usebox{\resBox}};

        \end{tikzpicture}
    }
    \vspace{-2em}
    \caption{Gait embedding network with \textit{semi-hard} mining used for the training process. The input is a data batch of $N$ tensors that hold the 3D position of $J$ joints over sequences of $T$ frames. The input is processed by the Siamese network which computes $D$-dimensional embedding vectors. Every embedding vector is matched with positive samples of the same person and negative samples of a different one to form triplets. The negative sample is selected by searching for embeddings close to the anchor vector to make the learning process more efficient. With the selected triplets, the loss is calculated and backpropagated through the network. After training, the network is able to map the same person close by and different ones apart.}
    \vspace{-.5em}
    \label{fig:network_h}
\end{figure*}
 
\subsection{Capture Space Setup}
To capture gait data from participants, 25 smart edge sensors with RGB-D cameras are deployed throughout a $\sim$\SI{240}{\square\meter} lab space, mounted at $\sim$\SI{2.5}{\meter} height~\cite{Bultmann_RSS_2021,bultmann_ias2022}.
Each smart edge sensor estimates the 2D poses of detected persons from the RGB image stream by inferring a heatmap to locate $J=19$ keypoints defined at the major joints of the human skeleton including nose, ears, and eyes.
The CNN inference for the person detection and pose estimation runs locally on each sensor board using lightweight model architectures and embedded inference accelerators~\cite{Bultmann_RSS_2021,bultmann_ias2022}.
Subsequently, only the estimated 2D pose coordinates are transmitted to a backend system, where multiple camera views are fused to estimate a 3D skeleton model for each observed person, as illustrated in \reffig{fig:3DModel}\,(a). Leveraging prior knowledge of human skeletal dimensions (e.g. bone lengths), the estimated pose coordinates are refined and then re-projected back to the sensor boards in a semantic feedback loop to further improve the accuracy of the pose estimation~\cite{Bultmann_RSS_2021,bultmann_ias2022}.

\subsection{Data Collection}
For data collection, 22 persons without pathological or neuronal gait disorders volunteered to participate in the experiment. They were instructed to walk naturally, as they would in their everyday environment, for a duration of one minute. During this time, their joint positions were tracked and recorded by the above-described smart edge sensor system. Further details on the participants are given in \reftab{tab:TestSubjects}.

\begin{table}[t]
	\centering
	\caption{Participants in the data collection.}
	\vspace{-.7em}
	\begin{tabular}{cccc}
		\toprule
		Age & Height [cm]  & \multicolumn{2}{c}{Gender} \\
		\midrule
		  30 $\pm$ 10 & 176 $\pm$ 8 & 19 Male & 3 Female \\
		\bottomrule
	\end{tabular}%
	\label{tab:TestSubjects}%
	\vspace{-1em}
\end{table}

The resulting gait data from 20 participants was used to train a neural network for gait sequence embeddings. The data of each subject was divided into a larger training set (90\%) and a smaller validation set (10\%). The gait data from the remaining two participants was completely excluded from the training and validation process and was used exclusively for testing the performance of the final model.

\subsection{Data Preprocessing}
After data collection, the resulting gait tracks were filtered to exclude parts where the camera system failed to detect all participant's joints, as noisy, incomplete observations could lead to unrealistically long or distorted limb representations.

The coordinates of the remaining gait tracks were normalized to be independent of the subject's height, absolute position in the capture space, and orientation. For this, the height of each person is scaled to $1$ and the root coordinate, located at the pelvis, is subtracted from each skeleton. Next, to normalize the orientation, the joint coordinates are rotated to align the $x$-axis with the connection of the left and right hip and the $z$-axis with the direction of the neck and pelvis (cf. Fig.~\ref{fig:3DModel}\,(b)). This ensures that all participants are constantly oriented as if walking straight forward, and the computed embeddings are independent of the absolute position and walking direction.

The resulting 3D joint coordinates for each frame are then represented as a matrix $X\in\mathbb{R}^{J\times3}$, with $J=19$ the number of tracked keypoints. For a gait sequence, the joint coordinates of a subject are then stacked into tensors with the shape of $(J, T, 3)$, illustrated in Fig.~\ref{fig:3DModel}\,(c). Each tensor includes the gait coordinates of a one-second walking sequence, resulting in a sequence length of $T=30$ at a \SI{30}{\hertz} capture frame rate. Finally, the tensors are matched with a label indicating the participant's identification number.

\subsection{Gait Embedding Network}
The network architecture used to identify individuals by their walking pattern is based on a TriNet Siamese network~\cite{hermans2017triplet} using a lightweight ResNet\,18~\cite{he_deep_2016} backbone for efficient computation.
We further employ a semi-hard mining strategy for efficient triplet selection during training. The network architecture and training process are illustrated in Fig.~\ref{fig:network_h}.
As input, the network receives data batches of the preprocessed and normalized gait sequences, each consisting of $N$ tensors with dimensions $(J, T, 3)$. The ResNet\,18 backbone maps each input tensor to a $(512, 1, 1)$ feature vector. This feature vector is flattened and projected to a $D$-dimensional embedding space using a fully connected linear layer. Finally, the embedding vector is $L_2$-normalized. The network outputs a data batch of $N$ $D$-dimensional gait sequence embedding vectors. As detailed in \refsec{sec:Evaluation}, we choose a batch size of $N=64$ and an embedding dimension of $D=32$ in our experiments. To enhance the network's performance, we initialized the ResNet\,18 model with ImageNet-pretrained weights.

\begin{figure}[t]
    \centering \footnotesize
    \begin{tabular}{cccc}
    \includegraphics[width=.95\linewidth, clip, trim={300 200 50 200}]{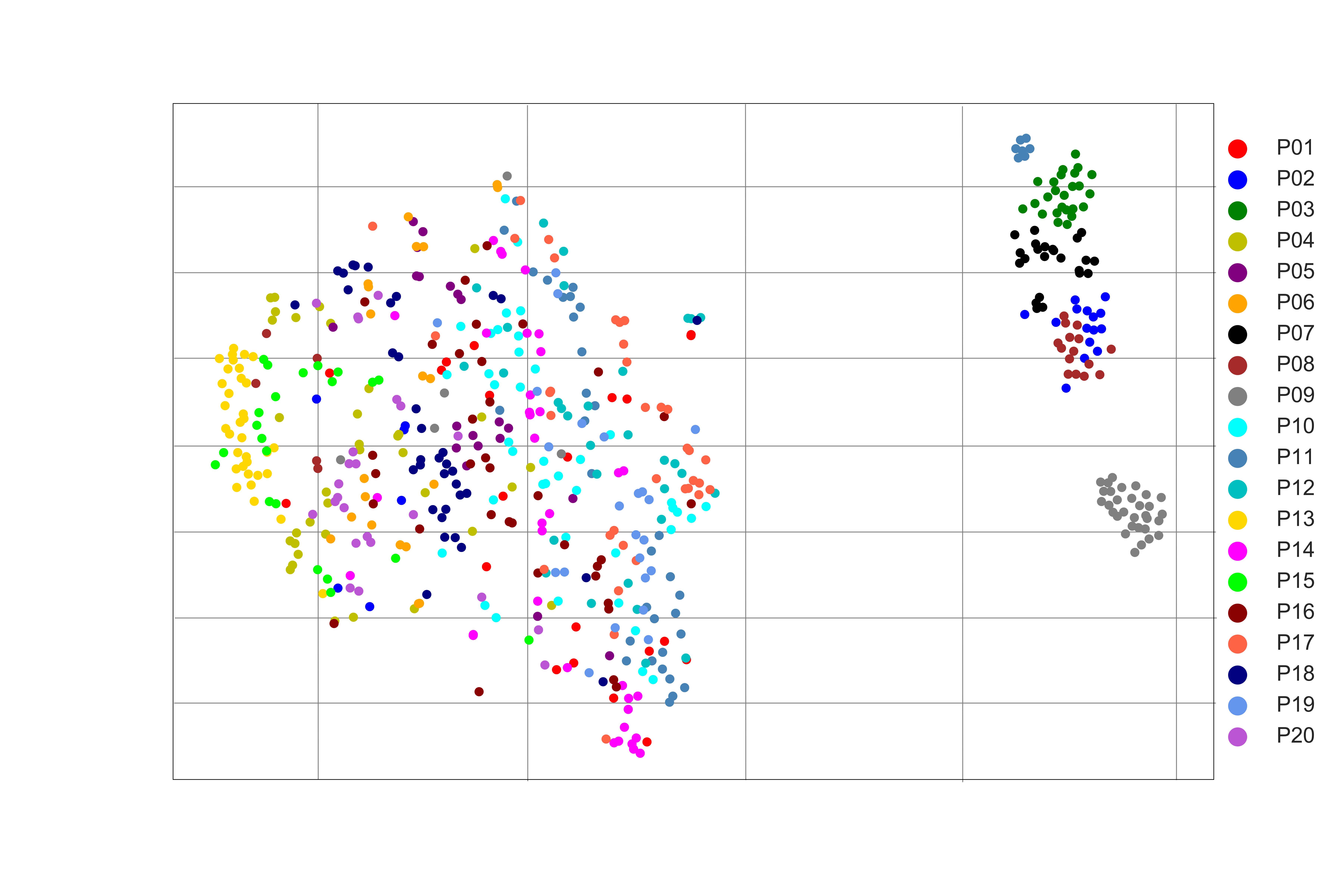}\vspace*{-0.5ex}\\
    \textbf{(a)} raw input data \\[6pt]
    \includegraphics[width=.95\linewidth, clip, trim={300 200 50 200}]{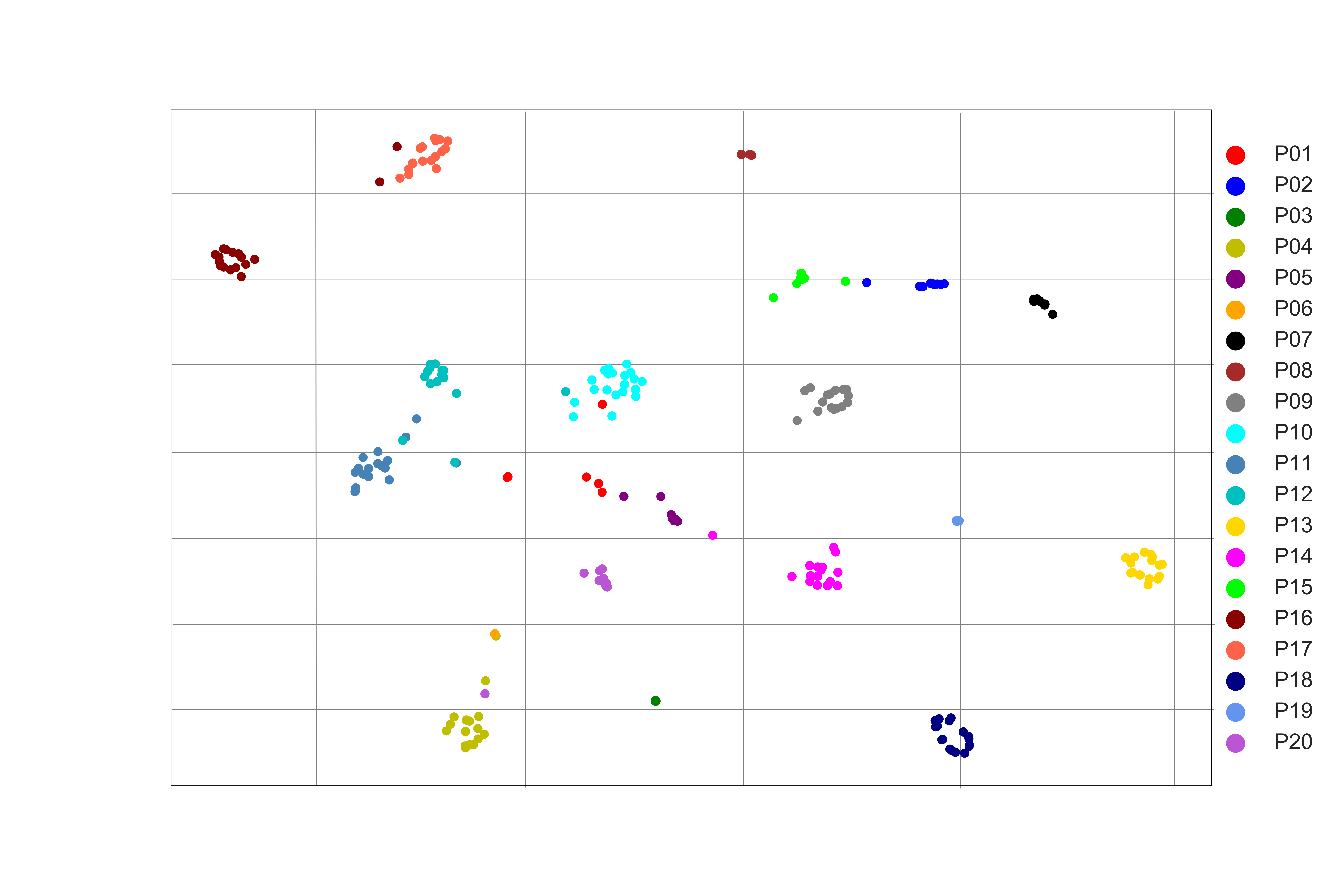}\\
    \textbf{(b)} final model  
    \end{tabular}
    \caption{t-SNE projections of the validation data. Each dot represents a validation sequence with Participant ID coded by color. (a) raw input data, (b) embeddings computed with final model after $10,000$ training epochs.}
    \label{fig:TSNE}
    \vspace{-1em}
\end{figure}

\subsection{Triplet Loss with Semi-hard Mining}
To enhance the speed and efficiency of the training process, we employ semi-hard negative mining for online triplet selection~\cite{schroff2015facenet}. In this process, first, a batch of input gait sequences is sampled from the training data and processed with the network to compute the embedding vectors.
Each output embedding then serves as a possible anchor to form triplets.
For this, the anchor is associated with a positive example, which is an embedding vector from the same person as the anchor, and a negative example.
The negative match is selected by calculating the loss values of all possible negative pairs, as shown in Eq.~(\ref{al:loss}):
\begin{align}
    \label{al:loss}
    \text{Loss} = \overline{\text{ap}} \ - \ \overline{\text{an}} \ + \ \text{margin}.
\end{align}

The Euclidean distance between the anchor and a negative example ($\overline{\text{an}}$) is subtracted from the distance between the anchor and the positive example ($\overline{\text{ap}}$). Finally, a pre-selected margin $\delta=0.2$, is added to this calculation.
After calculating the losses of every possible negative match in the data batch given a pair of anchor and positive example, triplets with a loss greater than zero, but less than the margin are selected. The negative sampling is repeated for all $\overline{\text{ap}}$ pairs.
This \textit{semi-hard} mining strategy~\cite{schroff2015facenet} selects negative samples that are further away from the anchor than the positive sample but still hard, as the embedding distance is close to the $\overline{\text{ap}}$ distance. In contrast, selecting only the hardest negative can lead to model collapses, as confirmed in \reftab{tab:Triplet selection}.

The model's loss is computed by averaging the losses of all found triplets, calculated as in Eq.~(\ref{al:loss}). This loss is then back-propagated through the network to adjust and improve its weights. The network's ultimate goal is to produce embedding vectors that have small distances when representing the same person by learning to differentiate their gait patterns.

\section{Evaluation}
\label{sec:Evaluation}
To evaluate the clustering ability of the proposed network, we apply the K-means algorithm to the embedding vectors computed from the validation data. We calculate the Adjusted Rand Index (ARI) score~\cite{hubert1985comparing} based on the resulting clusters for quantitative evaluation of the clustering accuracy. Further, we conduct ablation studies on the design choices of our network.
To illustrate the network's clustering ability, we employ the t-SNE projection~\cite{tsne} to display the similarities of the embedding vectors in a 2D scatter plot.
After completing the training process, the final network was additionally tested using the gait data of two previously unseen participants to assess the generalization capabilities of our system.

To evaluate our approach on activity clustering when dealing with larger data variety, we trained and tested the network using the popular Human~3.6M dataset~\cite{h36m}.

\begin{figure}[t]
    \centering
    \includegraphics[width=1\linewidth, clip, trim={100 80 10 80}]{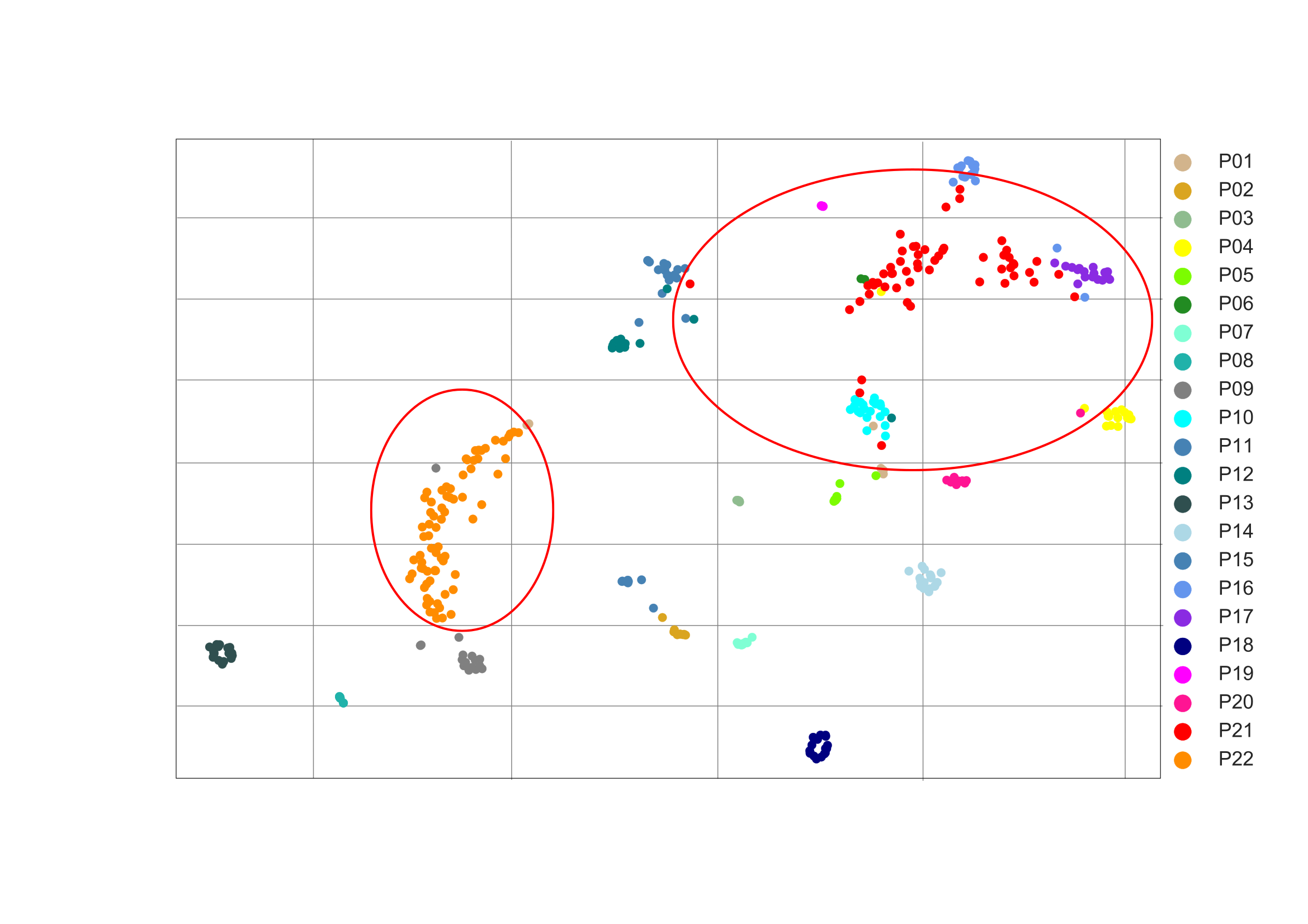}
    \vspace{-2em}
    \caption{t-SNE plot for validation of the model with gait data of two unknown participants 21 (red) and 22 (orange), highlighted with red circles.}
    \label{fig:TSNE_valid}
    \vspace{-1em}
\end{figure}

\begin{figure}[t]
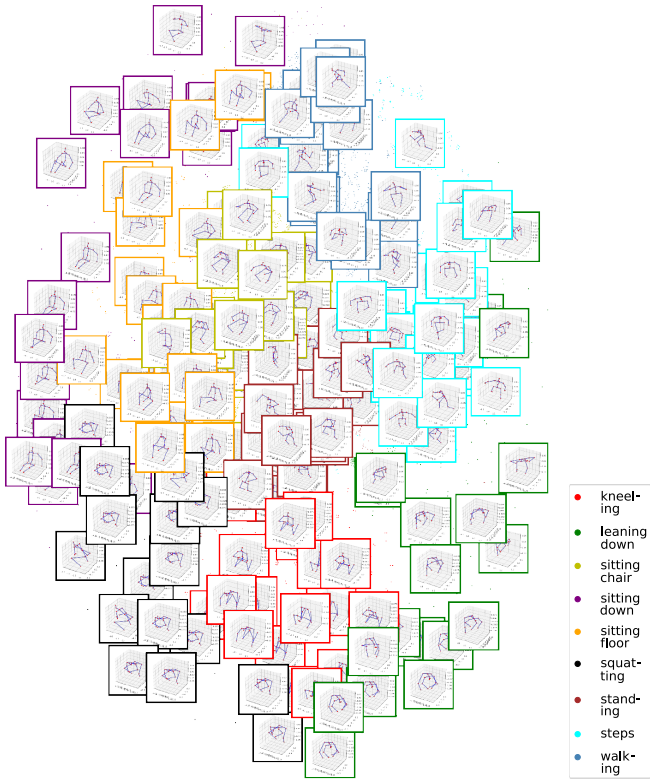

    \centering
    \includegraphics[height=10.25cm,trim= 0 10 0 25, clip]{images/Cluster_Human36M_bg.png}
		\includegraphics[height=3.9cm]{images/Cluster_Human36M_legend3.pdf}
    \vspace{-.5em}
    \caption{t-SNE plot of activities from Human 3.6M (zoom into \reffig{fig:Titel}\,(c)).}
    \label{fig:H3M}
\end{figure}

\subsection{Clustering Performance}

The t-SNE plot in Fig.~\ref{fig:TSNE} compares the direct clustering of the raw input data to the network's clustering performance. 
During training, the network improves its ability to recognize different gait patterns, learning to separate the participants into distinct clusters. In the raw data, Fig.~\ref{fig:TSNE}\,(a), two clusters of several participants are visible, but a clear separation of individual subjects is not possible. Using the gait embedding vectors of the trained network, Fig.~\ref{fig:TSNE}\,(b), nearly every participant has a separate cluster.

To evaluate the generalization performance of our network, we test it with gait data of two unknown subjects, completely unseen during the training process, as shown in Fig.~\ref{fig:TSNE_valid}.
The model successfully maps the unknown gait sequences close to each other, although they are slightly more spread out compared to the validation data of known participants. Notably, Participant 21 has several gait sequences incorrectly assigned to clusters of nearby participants.

\begin{figure}[t]
    \centering
    \includegraphics[width=1\linewidth]{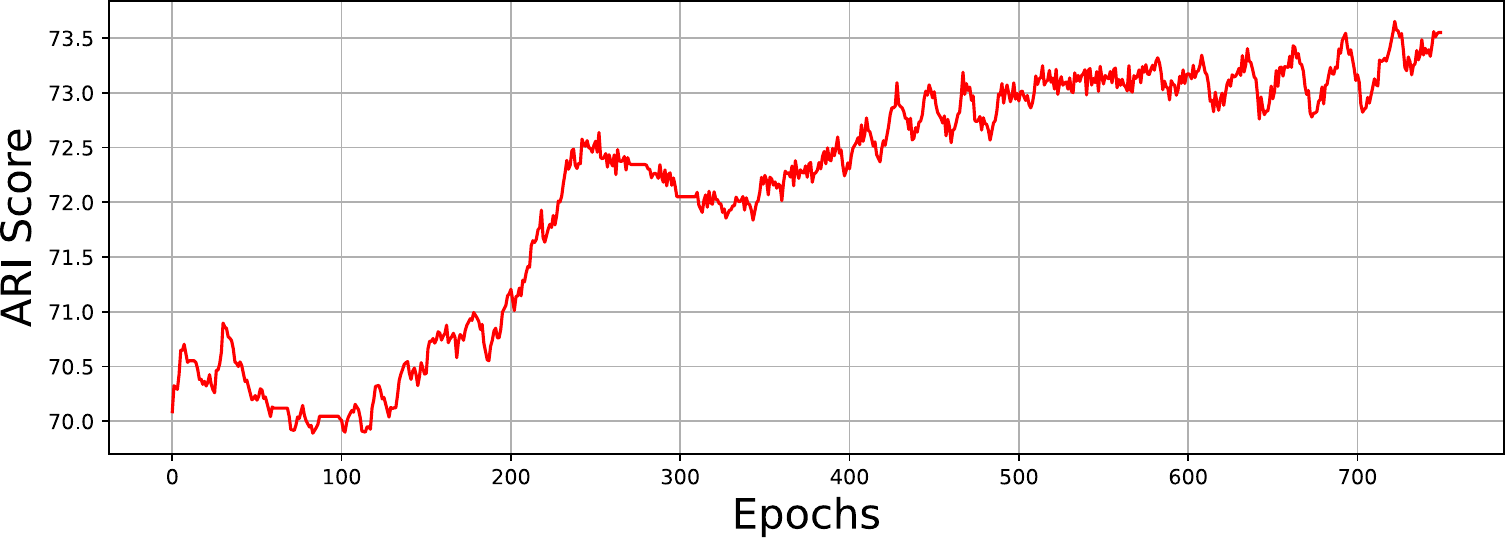}
     \vspace{-2.em}
    \caption{ARI score evolution during training for H3.6M activity clustering.}
    \label{fig:H36m_valid}
\end{figure}

Further experiments are performed with the public Human~3.6M database~\cite{h36m}, where sequences are clustered according to the activity labels of Tanke et al.~\cite{tanke2021intention}. The clustering is visualized in \reffig{fig:H3M}, showing the first frames of the validation sequences. Different actions are nicely separated and similar activities lie closely together in the embedding space, i.e. sitting to the top-left, walking or standing to the top-right, and kneeling or crouching to the bottom. The evolution of the ARI score during training is shown in \reffig{fig:H36m_valid}, with a final score of $73.5\%$ after 750 epochs.

\subsection{Ablation Studies}
The proposed neural network was initially trained with several different parameters to find the most efficient choices. The tested parameters include the triplet selection method, the size of the embedding dimension, the size of the data loader batches, and the number of frames in each movement sequence. Parameters like the learning rate, $\text{lr}=10^{-4}$, and the margin, $\delta=0.2$, were selected as recommended in \cite{schroff2015facenet}. Tables \ref{tab:Triplet selection}--\ref{tab:sequ_len} list the resulting ARI scores.

\begin{table}[t]
	\centering
	\caption{Clustering performance for triplet selection methods.}
	\vspace*{-1.5ex}
	\begin{tabular}{l|ccc|c}
		\toprule
		Method & random &  semi-hard mining & hard mining  & raw data\\
		\midrule
		 ARI score & 77.3\% &\textbf{80.3\%} & 3.4\% & 17.3\%\\
		 \bottomrule
	\end{tabular}%
	\label{tab:Triplet selection}%
\end{table}
\begin{table}[t]
	\centering
	\caption{Clustering perf. for embedding dimensions and batch sizes.}
	\vspace*{-1.5ex}
	\begin{tabular}{l|ccc|ccc}
		\toprule
		Emb. dim. & 32 & 64 & 128 & \multicolumn{3}{c}{32}\\
		\cmidrule(lr){2-4}\cmidrule(lr){5-7}
		Batch size & \multicolumn{3}{c|}{64}& 32 & 64 & 128\\
		\midrule
		ARI score & \textbf{80.3\%} & 80.0\% & 77.6\% & 79.1\% & \textbf{80.3\%} & 77.2\% \\
		\bottomrule
	\end{tabular}
	\label{tab:EmbDim}%
\end{table}
\begin{table}[t]
	\centering
	\caption{Cl. perf. for input sequence lengths and training iterations.}
	\vspace*{-1.5ex}
	\begin{tabular}{l|ccc}
		\toprule
		Seq. length & 15 & 30& 45 \\
		\midrule
		ARI score (1,000 epochs) & \textbf{80.3\%} & 78.9\% & 74.1\% \\[.3ex]
		ARI score (10,000 epochs) & 86.9\% & \textbf{87.8\%} & 84.6\% \\
		\bottomrule
	\end{tabular}%
	\label{tab:sequ_len}%
\end{table}%

Table~\ref{tab:Triplet selection} shows the model's clustering accuracy for different triplet selection procedures. \textit{Semi-hard} mining achieves the highest accuracy, with $80.3\%$ after 1,000 training epochs, while hard mining results in a model collapse with only $3.4\%$ accuracy. Using a random selection procedure, the accuracy is lower than with semi-hard mining, at $77.3\%$, still reaching a reasonable level of performance, significantly better than clustering the raw data directly ($17.3\%$).

Comparing the different embedding dimensions $D$ in Table~\ref{tab:EmbDim}, the resulting ARI scores are fairly similar.
With $D=32$, the model achieves an accuracy of $80.3\%$, $0.3\%$ higher than with $D=64$. Increasing the embedding dimension further ($D=128$) results in a decrease in ARI score to $77.6\%$. Additionally, training the model with larger embedding dimensions takes longer and requires more computational resources.
Modifying the batch sizes shows little impact on model performance, as apparent in Table \ref{tab:EmbDim}. The best result of $80.3\%$ is achieved with a batch size of 64. A batch size of 32 leads to an accuracy of $79.1\%$, while a batch size of 128 results in a lower accuracy of $77.2\%$.

Comparing the ARI scores after 1,000 and 10,000 epochs in Table~\ref{tab:sequ_len} reveals an improvement of approx. $10\%$ for sequence lengths $T$ of 30 and 45 frames, and $6.3\%$ for $T=15$. The highest accuracy, $87.8\%$, was achieved when the model completed 10,000 training epochs with a sequence length of 30 frames.
It is also noticeable that for a training time of 1,000 epochs, the accuracy decreases as the sequence length increases. However, when training for 10,000 epochs, the ARI score improves when the sequence length is doubled to $T=30$ frames but decreases when the length is increased again. Notably, all the scores after 10,000 epochs are higher than those after 1,000.
It seems plausible that the model requires more training time to recognize patterns when processing longer input sequences that give more context, leading to increased accuracy. Visualizing the participants during these sequences suggests that 15 frames roughly correspond to one step, depending on the participant's walking speed.
On the other hand, training the model with a sequence length of $T=45$ results in decreased accuracy. This may be due to the reduction in the total number of input tensors as the sequence length increases while the available data is limited. To determine whether a higher sequence length could further improve the model, additional gait measurements over longer periods could be beneficial.

For the final model, the input data loader uses a batch size of $N=64$, where each batch consists of gait sequences of length $T=30$ frames. The network outputs a $D=32$-dimensional embedding vector. During the training process, triplets were selected using the \textit{semi-hard} mining method, and the model was trained for 10,000 epochs.

\section{Conclusions}
\label{sec:Conclusion}
In this work, we proposed a novel approach to enable human gait clustering using a neural network to learn movement patterns and separate people by their gaits.
The gait data was recorded using a smart edge sensor network providing accurate marker-less motion capture.

We implemented a Siamese network based on TriNet~\cite{hermans2017triplet} with an ImageNet-pretrained ResNet-18 backbone and triplet loss calculation, employing semi-hard negative mining for efficient online triplet sampling. The network outputs 32-dimensional embedding vectors that cluster similar gait data while separating different patterns. Trained for 10,000 epochs on one-second gait sequences from 20 healthy individuals, it was tested with gait data from two new participants to validate its performance on unseen data.

While the clustering performance on raw gait data is poor, our model achieved an ARI score of 87.8\% with gait embedding vectors computed from 30-frame sequences, effectively differentiating all participants. Testing on unknown participants showed promising generalization, though with slightly less tight clustering than known subjects. We further demonstrated the network’s application in activity clustering using the Human3.6M dataset~\cite{h36m}.

Improving the dataset by including longer sequences and more participants could enhance performance, potentially requiring additional training iterations. A smaller, less cluttered measurement space could reduce joint displacement errors and unusable sequences. Including participants with diverse gait characteristics, such as elderly or injured, could further test and enhance the system’s robustness and applicability.

The smart edge sensor system offers a solution to perform instrumented gait analysis in a clinical setup. Because of its ability to recognize joints and capture movement without the need to apply sensors or markers, it significantly speeds up the process of performing gait analysis and improves the accessibility to digital gait data of patients. Nevertheless, before the system can be used for gait analysis in sports and medicine, its accuracy in detecting and tracking joints would need to be validated to the current gold standard of marker-based motion capture.

\bibliographystyle{IEEEtran}
\bibliography{literature}

\end{document}